\documentclass[graybox, a4paper]{svmult}

\usepackage{type1cm}        
\usepackage{makeidx}         
\usepackage{graphicx}        
\usepackage{multicol}        
\usepackage[bottom]{footmisc}
\usepackage{multirow}
\usepackage{diagbox}
\usepackage{amsmath,amssymb}
\usepackage{newtxtext}       %
\usepackage{newtxmath}       
\newcommand{\ourmodel}{Ours} 

\usepackage{color}
\usepackage{tikz}
\usepackage{hyperref}

\usepackage{mathrsfs}
\definecolor{dred}{rgb}{1,0,0}

\usepackage{xparse}

\tikzset{
  annotatedImage/x/.initial = 0.7,
  annotatedImage/y/.initial = 0.7,
  annotatedImage/width/.initial = 1,
  annotatedImage/.unknown/.code = {
    \edef\tikzappend{\noexpand\tikzset{annotatedImage/.append style =
                {\pgfkeyscurrentname=\pgfkeyscurrentvalue}}}
    \tikzappend
  },
  annotatedImage/.style = {
  }
}

\newsavebox\annotatedImageBox

\newcommand\AnnotatedImageVal[1]{\pgfkeysvalueof{/tikz/annotatedImage/#1}}
\newcommand\SetUpAnnotatedImage[2]{
    \tikzset{annotatedImage/.cd, #1}%
    \sbox\annotatedImageBox{\includegraphics[width=\AnnotatedImageVal{width}\textwidth,
                                          keepaspectratio]{#2}}%
    \pgfmathsetmacro\annotatedHeight{\ht\annotatedImageBox/28.453}
    \pgfmathsetmacro\annotatedWidth{\wd\annotatedImageBox/28.453}%
}

\NewDocumentCommand\annotatedImage{ O{} m m}{%
  \bgroup
    \SetUpAnnotatedImage{#1}{#2}%
    \begin{tikzpicture}[xscale=\annotatedWidth, yscale=\annotatedHeight]%
        \node[inner sep=0, anchor=south west] (image) at (0,0) {\usebox{\annotatedImageBox}};
        \node[annotatedImage] at (\AnnotatedImageVal{x},\AnnotatedImageVal{y}) {#3};
    \end{tikzpicture}
  \egroup%
}

\newcommand\annotate[1][]{\node[annotatedImage,#1]}

\newenvironment{AnnotatedImage}[2][1]{%
  \SetUpAnnotatedImage{#1}{#2}%
  \tikzpicture[xscale=\annotatedWidth, yscale=\annotatedHeight]
    \node[inner sep=0, anchor=south west,inner sep=0] at (0,0) {\usebox{\annotatedImageBox}};
}{\endtikzpicture}

\makeatletter
\newcommand\footnoteref[1]{\protected@xdef\@thefnmark{\ref{#1}}\@footnotemark}
\makeatother

\makeindex             

\begin{document}
\title*{Composing Pick-and-Place Tasks By Grounding Language}
%
\author{Oier Mees and Wolfram Burgard
\thanks{All authors are with the University of Freiburg, Germany. Wolfram Burgard is also with the Toyota Research Institute, USA. Corresponding author's email meeso@informatik.uni-freiburg.de 
}
}
%
%
\maketitle

\abstract{Controlling robots to perform tasks via natural language is one of the most challenging topics in human-robot interaction. In this work, we present a robot system that follows unconstrained language instructions to pick and place arbitrary objects and effectively resolves ambiguities through dialogues. Our approach infers objects and their relationships from input images and language expressions and can place objects in accordance with the spatial relations expressed by the user. 
Unlike previous approaches, we consider grounding not only for the picking but also for the placement of everyday objects from language. Specifically, by grounding objects and their spatial relations,  we allow  specification of complex placement instructions, e.g. ``place it behind the middle red bowl''. 
Our results obtained using a real-world PR2 robot demonstrate the effectiveness of our method in understanding pick-and-place language instructions and sequentially composing them to solve
 tabletop  manipulation tasks.
Videos are available at \url{http://speechrobot.cs.uni-freiburg.de}}

\section{Introduction}
\label{sec:1}
As robots become ubiquitous across human-centered environments the need for natural and effective human-robot communication grows. Natural language provides a rich and intuitive  way for humans
and robots to interact  due to the possibility of referring to abstract concepts. Moreover, many real-world tasks can be effectively described by  a series of language instructions. In this work, we aim to develop an approach that enables a robot to solve complex manipulation tasks by understanding a series of unconstrained language expressions characterizing pick-and-place commands. To do so, the robot has to locate unconstrained object categories based on arbitrary natural language expressions, known as referring expression comprehension, and understand spatial relations to generate object placing locations. In other words, the robot needs to ``ground'' the referred objects and their spatial relations from language in its world model. 
 
\begin{figure}[t]
    \centering
    \includegraphics[width=0.98\linewidth]{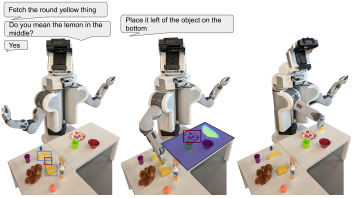}
    \caption{The goal of our work is to control a robot to perform tabletop manipulation tasks via natural language instructions. Our approach is able to segment objects in the scene, locate the objects referred to in language expressions, solve ambiguities through dialog and place objects in accordance with the spatial relations expressed by the user.}
    \label{fig:motivation}
\end{figure}

 However, understanding unconstrained language instructions is challenging due to the complexity and wide variety of abstract concepts expressed via human language, e.g. ``fetch the yellow thing'' and ``place it left of the bottom object''. Moreover, the expression might  contain ambiguities because there are several ``yellow things'' in which case the robot should be able to resolve the ambiguity through dialogue, as shown in Figure~\ref{fig:motivation}. Finally, the robot needs to reason about where to place the ``yellow thing'' relative to the ``leftmost container'' in order to reproduce the spatial relation ``right'', which is inherently ambiguous as natural language placement instructions do not uniquely identify a location in a scene. 

In this paper, we propose the first comprehensive system for controlling robots to perform tabletop manipulation tasks by sequentially composing unconstrained pick-and-place language instructions. Our approach  consists of two  neural networks. The first network learns to segment objects in a scene and to comprehend and generate referring expressions. The second network estimates pixelwise object placement probabilities for a set of spatial relations given an input image and a reference object. The interplay between both networks allows for an effective grounding of object semantics and their spatial relationships, without assuming a predefined set of object categories.
We demonstrate the effectiveness of our approach by enabling non-expert users to instruct tabletop manipulation tasks to a robot, based on sequences of pick-and-place speech commands. 
 
\section{Related Work}
Our work is primarily concerned with the task of grounding natural language instructions and spatial relations in the context of the robot's world model~\cite{clark1991grounding}.  Locating entities in images based on language is closely related to object recognition. Previous works in robotics~\cite{guadarrama2013grounding, pangercic2012semantic} have addressed semantic object retrieval by training classifiers to recognize predefined object categories. These approaches are limited in real-world scenarios as they are not capable of handling variation in the users natural language descriptions and are restricted to a small number of objects.

 Spatial relations also play a crucial role in understanding natural language instructions~\cite{hatori2018interactively, paul2016efficient}, as objects are often described in relation to others in tasks such as object placing~\cite{jiang2012learning, mees20icra_placements, mees17iros} or human robot interaction~\cite{guadarrama2013grounding, Shridhar-RSS-18, misra2016tell}. Concretely, spatial relations help the robot disambiguate multiple instances of the same object and to define target areas for placing the picked objects.
 In our previous work, we introduced a novel method to predict pixelwise object placement probability distributions for a set of commonly used prepositions in natural language~\cite{mees20icra_placements}. In contrast, we relax the assumption of having a single reference object on the tabletop and add a  grounding model to effectively place arbitrary objects in a scene that contains multiple objects.
 
 Recently, there has been significant progress made towards systems that can demonstrate their visual understanding by generating or responding to natural language in the context of images~\cite{kazemzadeh2014referitgame,antol2015vqa, johnson2016densecap, anderson2018vision,reed2016generative}. To learn joint visual-linguistic representations, state-of-the-art approaches use convolutional neural networks to encode visual features and recurrent neural networks to process language, replacing traditional handcrafted visual features and language parsers.  We leverage advances in modular networks~\cite{yu2018mattnet, hu2017modeling, andreas2016neural}  for  referential expression comprehension. This allows decomposing language into modular components related to subject appearance, location, and relationship to other objects, flexibly adapting to expressions containing different types of information in an end-to-end fashion.

Most related to our approach are the works by Shridhar \emph{et al.}~\cite{Shridhar-RSS-18} and Hatori \emph{et al.}~\cite{hatori2018interactively}, as both use an interactive fetching system to localize  objects mentioned in referring expressions with bounding boxes.  We tackle temporally more extended tasks, using our model which enables complex object placement commands such as ``place the cup on top of the leftmost box''. Notably, sequentially composing  pick-and-place language instructions can lead to desirable high-level behaviours, such as tidying up a tabletop or table setting for example.
Finally, in contrast to the template-based picking approaches of prior interactive fetching systems~\cite{Shridhar-RSS-18, hatori2018interactively} we leverage state-of-the-art methods for grasping  novel objects with 6-DOF grasps~\cite{gualtieri2016high}. 

\section{Method Description}
\begin{figure}[t]
    \centering
    \includegraphics[width=0.99\linewidth]{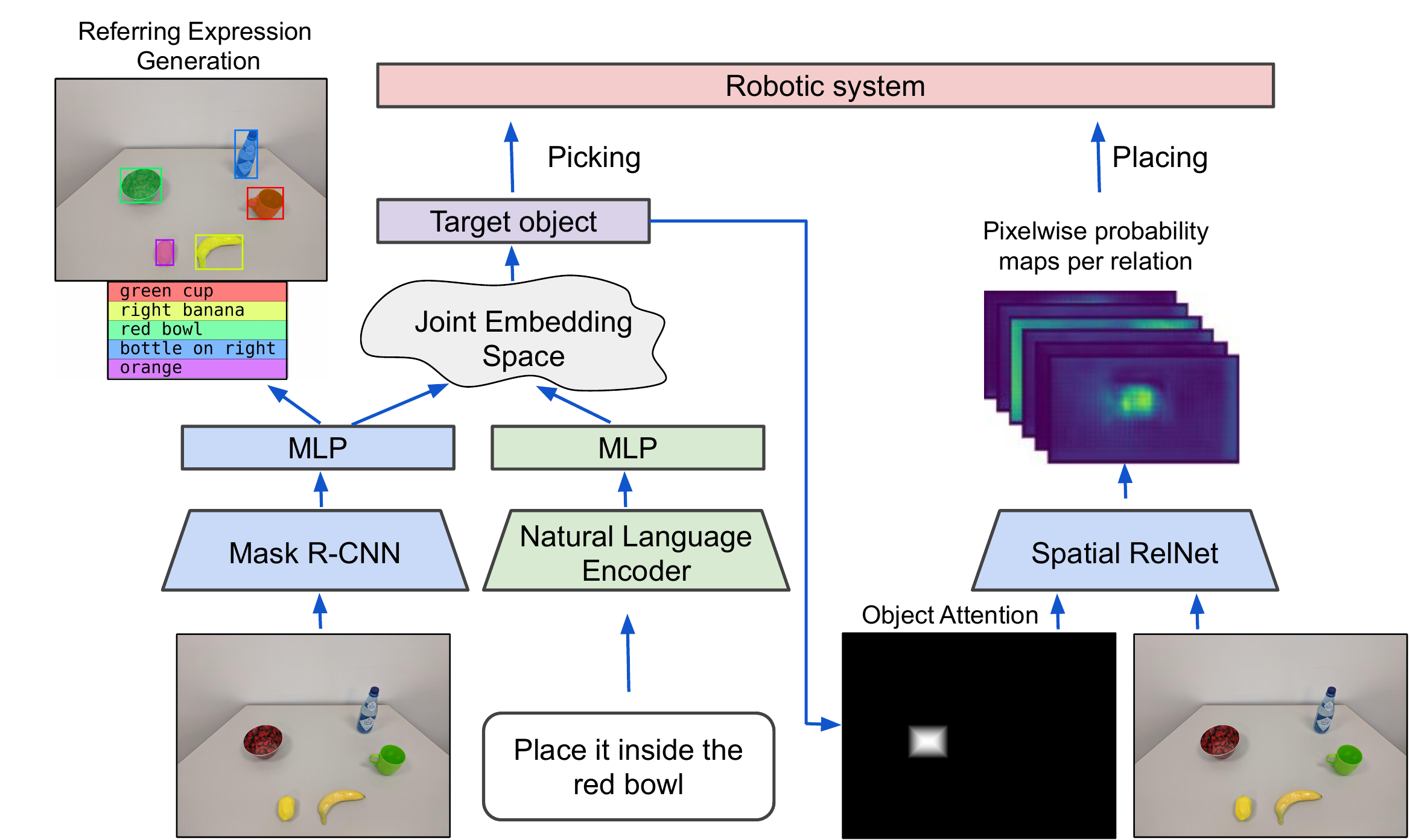}
    \caption{Overview of the system architecture. Our grounding network processes the input sentence and visual object candidates detected with Mask-RCNN~\cite{he2017mask} and performs referential expression comprehension. Additionally, it generates referential expressions for each object candidate to disambiguate unclear instructions. Once the reference object of a relative placement instruction has been identified, a second network  predicts object placing locations for a set of spatial relations.}
    \label{fig:architecture}
\end{figure}

In this section we describe the technical details of our method to control a robot to perform tabletop manipulation tasks via natural language instructions. 
Our approach relies on two models: a grounding model that identifies the most likely object referred by a language instruction and a neural network that predicts object placing locations conditioned on a set spatial relation. An overview of the system is given in Figure \ref{fig:architecture}.

\subsection{Target Object Selection}
We start off by detecting and segmenting all objects in the scene. We train a semantic segmentation network based on Mask-RCNN~\cite{he2017mask} with a Resnet-101 backbone, which extracts a set of region proposals or object candidates $o_i$ from an image.
After all objects on the scene are recognized, we need to identify which object the user is referring to in its language instruction. Given an input image $I$  and expression $r$, the target object selection is formulated as a task to find the best bounding box from the set of predicted candidate boxes $O = \{o_i\}_{i=1}^N$.
Our grounding model is based on MAttNet~\cite{yu2018mattnet}, a modular referring expression comprehension network. To enable human-robot communication in cases of ambiguous instructions, we have extended it to support the generation of self-referential expressions, described in Section \ref{sec:caption}. 

The candidate regions are encoded by a neural network consisting of three modular grounding components related to subject appearance, location and relationship to other objects. These modules combine image features encoded by a Resnet-101 network with relational and geometric features pertaining to the neighborhood or context of each candidate region.
The language expression $r$ is encoded in a word embedding layer, which encodes each word in the input sentence to a vector representation, followed by a bi-directional Long Short-Term Memory (LSTM)  and a fully-connected (FC) layer. Additionally, the language network learns two types of attention: attention weights that are computed on each word for each module and are summarized as phrase embedding $q^m \mid m \in \{\text{subj, loc, rel}\}$, and module weights  $ [  w_{subj}, w_{loc}, w_{rel} ] $ that estimate how much each module contributes to the overall expression score. 
Each visual module computes scores for each object candidate by calculating the cosine similarity between the vector representation of the instruction, and that of the candidate image region. Finally, the output module takes a weighted average of these scores to get an overall matching score $S(o_i \mid r) = w_{subj} S(o_i \mid q^{subj}) +  w_{loc} S(o_i \mid q^{loc}) +  w_{rel} S(o_i \mid q^{rel})$.  During training, we sample triplets consisting of a positive match $(o_i, r_i)$ and two random negative samples  $(o_i, r_j)$ and  $(o_k, r_i)$, where $o_k$ is some other object and $r_j$ is an expression describing some other object in the same image to apply a hinge loss:
\begin{equation}
\begin{split}
    \mathscr{L}_1  =&  \sum_i [ \lambda_1 \max (0, m_1 + S(o_i \mid r_j) - S(o_i \mid r_i))  \\
        & + \lambda_2 \max (0, m_1 + S(o_k \mid r_i) - S(o_i \mid r_i)) ].
\end{split}
\end{equation}
\subsection{Resolving Ambiguities}\label{sec:caption}
If the referred object cannot be uniquely identified by the grounding model, the system needs to ask for clarification from the human operator.  Inspired by recent advances in image caption generation and understanding~\cite{johnson2016densecap, mao2016generation, yu2017joint}, we incorporate a LSTM based captioning module to our grounding network that allows the robot to describe each detected object with a natural language description. Our referring expression generation module is jointly trained with our grounding network and shares the features used in the three modules related to subject appearance, location and relationship to other objects. 
Specifically, the visual target object representation $v_i^{vis}$ is modeled by a concatenation of the ResNet-101 C3 and C4 features, followed by one FC layer which is shared with the comprehension network  and one exclusive FC layer. To facilitate the generation of referential expressions that contain location information, such as  ``the cup in the middle'', we leverage the representation learned by the location module $v_i^{loc}$. This module combines a 5-d vector representing the top-left position, bottom-right position and relative area to the image for the candidate object, together with a relative location encoding of up to five surrounding objects of the same category. Finally, we integrate the output of the relationship module $v_i^{rel}$, which encodes the appearance and localization offsets of up to five category-agnostic objects in the targets surroundings to enable modeling sentences such as ``the teddy bear on top of the box''. 
The final visual representation for the target object is then a concatenation of the above features $v_i = [v_i^{vis}, v_i^{loc}, v_i^{rel}]$. The model is trained to generate sentences $r_i$ by minimizing the negative log-likelihood:
\begin{equation}
\mathscr{L}_2 = - \sum_i \log P (r_i \mid v_i).
\end{equation}
To generate discriminative sentences, we use a Maximal Mutual Information constraint proposed by Mao \emph{et al.}~\cite{mao2016generation} that encourages the generated expression to describe the target object better than the other objects within the image. Concretely, given a positive match $(o_i, r_i)$ we sample a negative  $(o_k, r_i)$,  where $o_k$ is some other object, and optimize the following max-margin loss:
\begin{equation}
\mathscr{L}_3 =  \sum_i [ \lambda_3 \max (0, m_2 + \log P (r_i \mid v_k) - \log P (r_i \mid v_i)).
\end{equation}
In order to detect if an instruction is ambiguous, we leverage the max-margin loss the comprehension model is trained with. Concretely, during training the  max-margin loss aims to guarantee that every correct pair of a sentence and an object has scores by a margin $m_1$ than any other pair with a wrong object or sentence. Therefore, if at test time there are more than one objects within that threshold, we consider them potential targets. For each candidate we generate multiple self-referential expressions via beam search and use the comprehension module to rerank these expressions and select the least ambiguous expression, similar to Yu \emph{et al.}~\cite{yu2017joint}. We then let the system ask the human ``Do you mean ...?''. After asking the question, the user can respond ``yes'' to choose the referred object or ``no'' to continue iterating through other possible objects. Alternatively, the user can provide a specific correcting response to the question, e.g., ``no, the banana on the right'', in which case we re-run our grounding module. 
\subsection{Relational Object Placement}
\begin{figure}[t]
\centering
\begin{AnnotatedImage}[width=1]{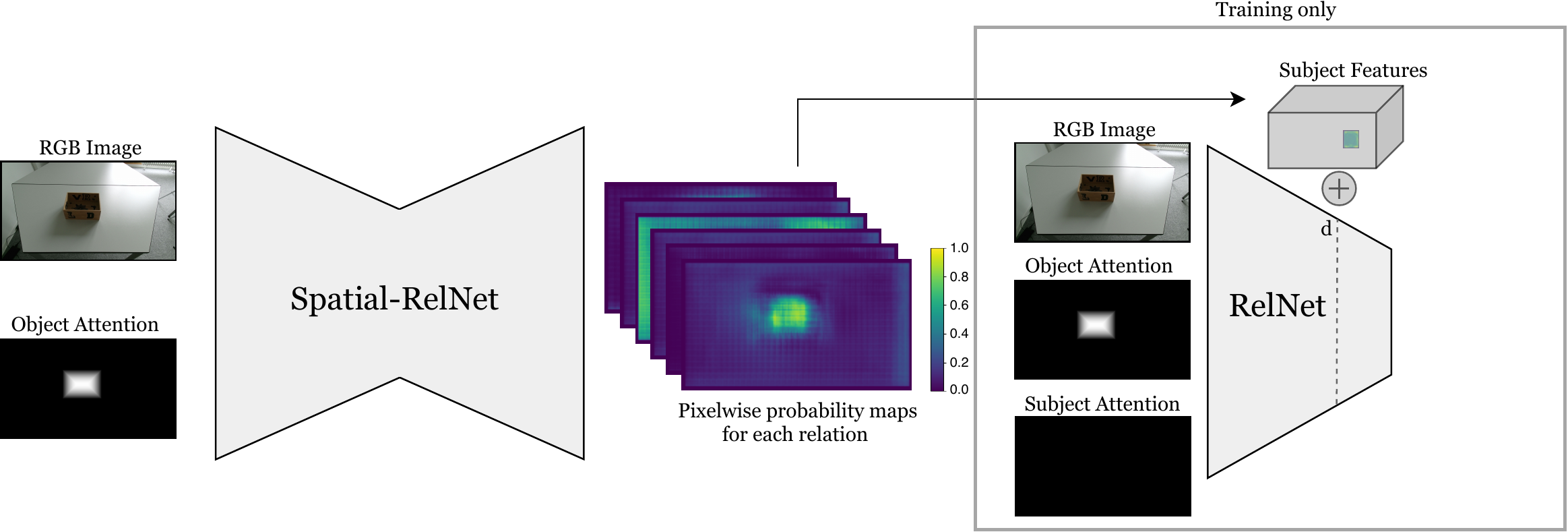}
\annotate (G) at (0.943,0.42){$ \scriptscriptstyle \left\lVert \Gamma(u,v) -  f_{\varphi} \right\rVert_2^2$};
\annotate (G) at (0.57,0.18){$\scriptscriptstyle(\Gamma)$};
\annotate (G) at (0.72,0.85){\tiny Sample $(u, v)$};
\end{AnnotatedImage}
  \caption{Our Spatial-RelNet~\cite{mees20icra_placements} network processes the input RGB image and an object attention mask to produce pixelwise probability maps $\Gamma$ over a set of spatial relations. During training, we sample locations $(u, v)$ according to $\Gamma$, implant inside an auxiliary classifier network at the sampled locations high level features of objects and classify the hallucinated scene representation to get a learning signal for Spatial-RelNet. At test time the auxiliary network is not used.}
\label{fig:spatial_architecture}
\end{figure}

Once an object has been picked, our system needs to be able to place it in accordance with the instructions from the human operator.  We combine referring expression comprehension with the grounding of spatial relations to enable complex object placement commands such as ``place the ball inside the left box''. Given an input image $I$ of the scene and the location of the reference item, identified with our aforementioned grounding module, we generate pixelwise  object placement probabilities for a set of spatial relations by leveraging the Spatial-RelNet architecture we introduced in our previous work~\cite{mees20icra_placements}. We consider pairwise relations and express the subject item as being \emph{in relation to} the reference item. We model relations for a set of commonly used natural language spatial prepositions $C= \{\texttt{inside}, \texttt{left}, \texttt{right}, \texttt{in}~\texttt{front}, \texttt{behind},  \texttt{on}~\texttt{top}\} $. 
As natural language  placement  instructions  do  not  uniquely  identify a  location  in  a  scene,  Spatial-RelNet predicts non-parametric distributions to capture the inherent ambiguity. A key challenge to learning such pixelwise spatial distributions is the lack of ground-truth data. Spatial-RelNet overcomes this problem by leveraging a novel auxiliary learning formulation, as shown in Figure~\ref{fig:spatial_architecture}. During training, pixel locations $(u,v)$  are sampled according to the probability maps $\Gamma$ produced by Spatial-RelNet. To get the learning signal, high level features of objects are implanted into a pretrained auxiliary classifier $f_{\varphi}$ to compute a posterior class probability over relations. This way, we can reason over what relation would most likely be formed if we placed an object at the given location.

\section{System Implementation}
\subsection{Machine Learning Setup} 
 During training, we sample the same triplets for both the object comprehension module and the expression generation module. We set the margin $m_1=0.1$ for the comprehension ranking and $m_2=1.0$ for the generation loss. We additionally use MAttNet's auxiliary visual attribute  classification loss.
 We use the Adam optimizer to train the joint model with an initial learning rate of 0.0004. For the contrastive pairs, we set $\lambda_1 = 1$, $\lambda_2 = 1$ and$\lambda_3=0.1$. We make the word embedding of the comprehension and generation modules shared to reduce the number of parameters.
 For implementation details of Spatial-RelNet, we refer  to the original paper~\cite{mees20icra_placements}.

\subsection{Robot Setup}
To pick an object from language, we first identify the object with our grounding model and extract the corresponding segmentation mask of the selected object.  We use an Amazon Echo Dot device to synthesize the voice instructions.
We  localize the object in 3D space and generate grasp poses with Grasp Pose Detection (GPD)~\cite{gualtieri2016high}, which predicts a series of 6-DOF candidate grasp poses given a 3D point cloud for a 2-finger grasp. The reachability of the proposed candidate grasps are checked using MoveIt!, and the highest quality reachable grasp is executed with the PR2 robot. For placing the object, we first sample a location from the spatial distribution predicted by our Spatial-RelNet model. We rely on keyword spotting to select the corresponding predicted distribution. Next,  we localize the pixel coordinate in 3D space and plan a top-down grasp pose to the calculated 3D point. Finally, the end-effector is moved above the desired location and then the gripper is opened to complete the placement.

\section{Experiments}
We evaluate our approach under two settings. First, we evaluate the capability of our approach to comprehend and generate referring expressions for a wide variety of objects on the RefCOCO dataset~\cite{kazemzadeh2014referitgame}. Next, we evaluate the ability of our robotics system to follow pick-and-place language instructions in human-robot experiments.
\subsection{RefCOCO Benchmark}
The RefCOCO dataset contains images and corresponding referring expressions that uniquely identify a wide variety of objects in the images. We compare our grounding networks ability to comprehend and generate referring expressions against several strong baselines on Table~\ref{tab:comprehension}. For evaluating the comprehension, we compute the intersection-over-union (IoU) of the selected region with the ground-truth bounding box, considering IoU $>0.5$ a correct comprehension. To evaluate the generation module, we leverage standard machine translation metrics commonly used in image captioning, such as METEOR and CIDEr. We observe that by jointly training the comprehension and language generation modules, they regularize each other and improve their respective performances, demonstrating the effectiveness of multitask learning~\cite{caruana1997multitask, yu2017joint, mees20icra_asn}.
\begin{table}[h]
  \centering
  \setlength{\tabcolsep}{5pt}
  \begin{tabular}{ |c|c|c|c|c|c|c|c|}
\hline
 & \multicolumn{3}{c|}{RefCOCO comprehension}  & \multicolumn{4}{c|}{RefCOCO generation}\\ 
\cline{2-8}
& val & TestA & TestB  & \multicolumn{2}{c|}{TestA} & \multicolumn{2}{c|}{TestB}\\ 
\cline{5-8}
&  & & & Meteor & CIDEr & Meteor & CIDEr \\
\hline
Mao~\cite{mao2016generation} & - & 63.15 & 64.21 & - & - &- &- \\ 
INGRESS~\cite{Shridhar-RSS-18} &  77  & 76.7 & 77.7 & - & - &- &- \\ 
SLR~\cite{yu2017joint, hatori2018interactively} & 79.56 & 78.95 & 80.22 & 0.268 & 0.697 & 0.329 & 1.323\\
MAttNet~\cite{yu2018mattnet} &  85.65  & 85.26 & 84.57  & - & - &- &- \\ 
\ourmodel{} & \textbf{86.15} & \textbf{87.18} & \textbf{85.36}& \textbf{0.29} & \textbf{0.753} & \textbf{0.33} & \textbf{1.33} \\ 
\hline
  \end{tabular}
  \caption{Referring expression comprehension and generation on the RefCOCO dataset, with human-annotated ground-truth object regions.}
  \label{tab:comprehension}
\end{table}
\subsection{Robot Experiments}
We evaluate our approach on two real-world scenarios: picking and placing objects according to user defined object arrangements and a tidy-up task. We will first describe the setup of the object arrangement experiment. Our study involved 4 participants recruited from a university community\footnote{Further quantitative experiments were infeasible at time of submission
due to COVID-19.}. The robots workspace contained two tables, as shown in Figure \ref{fig:motivation}. One table contained previously unseen objects in clutter. The second table contained a single reference object.  The average number of objects on the cluttered table was 5.6. The participants were asked to instruct a PR2 robot to arrange a desired target scene by picking objects from the cluttered table and using relational expressions to place them on the second table.  In addition to the robots RGB-D camera we placed a second camera in front of the cluttered table and performed online registration to compute a global point cloud. The tidy-up task consisted of iteratively picking 4 colored objects from the cluttered table and placing the same colored objects on the left container and the remaining objects on the right container. In this experiment we were interested in evaluating the number of actions the robot has to take to complete the task, given unambiguous instructions.

Table~\ref{tab:roboexp} shows the performance of our approach on a PR2 robot for the first experiment. Our approach achieves a 78.3\%  target object selection accuracy and a 85.7\% accuracy on selecting the reference object the placing will be relative to. 
\begin{table}[h]
  \centering
  \begin{tabular}{ c c c c c c c}
  & Target Object & Target Object & Placing Base & Placing & Avg. Number & Pick and\\
  &  Selection & Grasping & Grounding & Success & of Feedback &  Place\\
  \hline
  Ours & 78.3\% (47/60) & 74.4\% (35/47) & 85.7\% (30/35) & 83.3\% (25/30) & 0.63 (60/95) &  63\% (60/95)\\
  \end{tabular}
  \caption{Performance of our approach on a real robot platform following natural language instructions  to pick and place objects in a tabletop scenario.}
  \label{tab:roboexp}
\end{table}
The higher accuracy of the latter is due to fewer candidate objects being on the placing table and the participants preferring to use ambiguous expressions for the picking instructions. The robot took $\sim20$ seconds to complete an action from the moment the human started to speak. We report a grasping performance of 74.4\% with GPD. We find that some objects such as mugs are particularly difficult for GPD as it often fails to find feasible grasps due to either occluded object parts or noisy measurements on thin structures such as rims. Our object placement approach achieves a success rate of 85.7\%. We observe some failure cases for large object placements, because of missing 3D priors of the objects to be placed. Thus, when placing a big box left of a small box, it is possible that the chosen placement results in the big box partially ending up on top of the small box. For the tidy up task, we report a mean task length of 14.4 actions, due to several re-grasp attempts.
Overall, our results demonstrate the ability of our approach to allow non-expert users to instruct tabletop manipulation tasks based on sequences of  pick-and-place speech commands.

\section{Conclusions and Discussion} 
\label{sec:conclusion}
In this paper, we proposed the first robotic system that allows non-expert users to instruct tabletop manipulation tasks by sequentially composing unconstrained pick-and-place language instructions and can clarify a human operator's intention through  dialogue.
We demonstrate the effectiveness of our approach to encode high-level behaviours in a highly challenging, realistic environment. Even though we are far from achieving robots that can learn to relate human language to their world model, we hope our work is a step in this direction. 

While the experimental results are  promising, our approach has several limitations. First, relative object placement instructions do not allow for fine-grained target specification due to its inherent ambiguity.  Addressing this issue would require learning user preferences from feedback~\cite{abdo2016organizing}. Second, we observe some failure cases for large object placements, because of missing 3D priors of the objects to be placed. Integrating 3D priors is a natural extension to enable optimizing placement poses~\cite{haustein2019object} and to reason over the effects of actions on the scene~\cite{nematoli20iros}. Third, we find that GPD often fails to find feasible grasps due to either occluded object parts or noisy measurements. Integrating methods that can complete occluded scene regions~\cite{mees19iros, varley2017shape} or generate more diverse grasps~\cite{mousavian2019} might help alleviating these problems.
Finally, our approach is limited to tabletop tasks that can be characterized by pick-and-place actions. An exciting area for future work may be one that not only grounds object semantics and spatial relations, but also grounds actions in order to learn complex behaviours with language conditioned continuous control policies~\cite{lynch2020grounding, shaoconcept2robot}.

\section*{Acknowledgments}
This work has  been supported partly by the Freiburg Graduate School of Robotics and the German Federal Ministry of Education and Research under contract number 01IS18040B-OML. We thank Henrich Kolkhorst for his contributions to the speech-to-text pipeline and to Andreas Eitel for valuable discussions.

\bibliographystyle{unsrt}
\bibliography{references}

\end{document}